\documentclass[lettersize,journal]{IEEEtran}

\usepackage{amsmath,amsfonts}
\usepackage{helvet}  
\usepackage{courier}  
\usepackage[hyphens]{url}  
\usepackage{array}
\usepackage[caption=false,font=normalsize,labelfont=sf,textfont=sf]{subfig}
\usepackage{textcomp}
\usepackage{stfloats}
\usepackage{url}
\usepackage{verbatim}
\usepackage{graphicx}
\usepackage{cite}
\usepackage[ruled]{algorithm2e}
\usepackage{algpseudocode}
\usepackage{multirow}
\usepackage{booktabs}
\usepackage{color, soul} 
\usepackage {hyperref}
\usepackage{wrapfig}
\usepackage{soul}
\usepackage{makecell}
\usepackage[utf8]{inputenc}
\usepackage{threeparttable} 

\begin{document}

\title{Developmental Plasticity-inspired Adaptive Pruning for Deep Spiking and Artificial Neural Networks}

\author{Bing Han, Feifei Zhao, Yi~Zeng, Guobin Shen
\IEEEcompsocitemizethanks{
\IEEEcompsocthanksitem Bing Han is with the Brain-inspired Cognitive Intelligence Lab, Institute of Automation, Chinese Academy of Sciences, Beijing 100190, China, and School of Artificial Intelligence, University of Chinese Academy of Sciences, Beijing 100049, China.\protect\\
\IEEEcompsocthanksitem Feifei Zhao is with the Brain-inspired Cognitive Intelligence Lab, Institute of Automation, Chinese Academy of Sciences, Beijing 100190, China.\protect\\
\IEEEcompsocthanksitem Yi Zeng is with the Brain-inspired Cognitive Intelligence Lab, Institute of Automation, Chinese Academy of Sciences, Beijing 100190, China, and Center for Long-term Artificial Intelligence, Beijing 100190, China, and University of Chinese Academy of Sciences, Beijing 100049, China, and Key Laboratory of Brain Cognition and Brain-inspired Intelligence Technology, Chinese Academy of Sciences, Shanghai, 200031, China. \protect\\
\IEEEcompsocthanksitem Guobin Shen is with the Brain-inspired Cognitive Intelligence Lab, Institute of Automation, Chinese Academy of Sciences, Beijing 100190, China, and School of Future Technology, University of Chinese Academy of Sciences, Beijing 100049, China.\protect\\
\IEEEcompsocthanksitem The first and the second authors contributed equally to this work, and serve as co-first authors.\protect\\
\IEEEcompsocthanksitem The corresponding author is Yi Zeng (e-mail: yi.zeng@ia.ac.cn). \protect\\}
\thanks{This paper was produced by the IEEE Publication Technology Group. They are in Piscataway, NJ.}
\thanks{Manuscript received April 19, 2021; revised August 16, 2021.}}

\markboth{Journal of \LaTeX\ Class Files,~Vol.~14, No.~8, August~2021}%
{Shell \MakeLowercase{\textit{et al.}}: A Sample Article Using IEEEtran.cls for IEEE Journals}


\maketitle

\begin{abstract}
Developmental plasticity plays a prominent role in shaping the brain's structure during ongoing learning in response to dynamically changing environments. However, the existing network compression methods for deep artificial neural networks (ANNs) and spiking neural networks (SNNs) draw little inspiration from brain's developmental plasticity mechanisms, thus limiting their ability to learn efficiently, rapidly, and accurately. This paper proposed a developmental plasticity-inspired adaptive pruning (DPAP) method, with inspiration from the adaptive developmental pruning of dendritic spines, synapses, and neurons according to the ``use it or lose it, gradually decay" principle. The proposed DPAP model considers multiple biologically realistic mechanisms (such as dendritic spine dynamic plasticity, activity-dependent neural spiking trace, and local synaptic plasticity), with additional adaptive pruning strategy, so that the network structure can be dynamically optimized during learning without any pre-training and retraining. Extensive comparative experiments show consistent and remarkable performance and speed boost with the extremely compressed networks on a diverse set of benchmark tasks for deep ANNs and SNNs, especially the spatio-temporal joint pruning of SNNs in neuromorphic datasets. This work explores how developmental plasticity enables complex deep networks to gradually evolve into brain-like efficient and compact structures, eventually achieving state-of-the-art (SOTA) performance for biologically realistic SNNs. 
\end{abstract}

\begin{IEEEkeywords}
Integrated Development Mechanisms, Dendritic Spine Dynamic Plasticity, Activity-dependent Synaptic plasticity, Brain-inspired Deep ANNs and SNNs Compression
\end{IEEEkeywords}

\section{Introduction}
\IEEEPARstart{T}{he} human brain with highly plastic is the product of hundreds of millions of years of evolution, thereby allowing infants to emerge with high-level intelligence as they grow and develop. Neural circuits and network topology of the brain are also the products of development over an individual’s life span. Since the birth of the baby, synapses first undergo explosive growth, peaking by age two or three. Then those surplus synapses are gradually eliminated throughout childhood and adolescence according to adaptive pruning mechanisms\cite{huttenlocher2009neural,huttenlocher1979synaptic}. This developmental process dynamically shapes the network structure as a result of continuous interaction with the environment and neural changes induced by learning\cite{elman1996rethinking,johnson2001functional}. The developmental plasticity of the brain enables it to show remarkable plasticity in response to changing environments and to perform multiple complex cognitive functions with extremely low energy consumption. However, current deep neural networks (DNNs) and deep spiking neural networks (DSNNs) employ complex networks with a large number of parameters to solve a single task, which leads to prohibitively expensive computational costs and storage overhead. Besides, DNNs and DSNNs without developmental structural plasticity lack sufficient adaptability and flexibility in learning different tasks. This is a significant gap between baby-like highly-efficient learning and adaptive development.

Taking inspiration from the multi-scale developmental plasticity in the brain, whereby dendritic spines, synapses, and neurons adaptive formation and elimination according to the ``use it or lose it" principle, we proposed a generalized developmental plasticity-inspired adaptive pruning (DPAP) method for DNNs and SNNs. Incorporating DPAP into the ongoing learning and optimization of neural networks enables dynamically pruning redundant synapses and neurons according to their activity levels. Especially for event-driven SNNs, we use neuronal temporal spiking sequences to represent activity levels and incorporate temporal dimension pruning. Different from the existing network compression models, DPAP is remarkable at multiple levels, more biologically plausible with ongoing developmental plasticity, more adaptive pruning for efficient structure shaping, and naturally brings superior performance and learning speed.

The existing model compression methods are intended to reduce memory and operations consumption while minimizing accuracy drop. The DNNs compression methods include pruning\cite{han2015learning}, quantization\cite{han2015deep,hubara2017quantized}and knowledge distillation\cite{hinton2015distilling}. Here, we focus on the pruning methods. DNNs pruning methods considered weight magnitude\cite{molchanov2016pruning},weight gradient\cite{hassibi1993optimal, huang2018data}, weight similarity\cite{srinivas2015data},  Batch Normalization (BN) factor\cite{you2019gate} as evaluation criteria, pruning fine-grained individual parameters\cite{yang2018efficient} or coarse-grained overall structures\cite{wen2016learning,yu2022width}. Deep networks are very sensitive to such pruning strategies, thus pre-training and retraining are required to guarantee performance, which is not biologically plausible.  Some developmental plasticity-inspired pruning methods prune neurons or synapses adaptively through a biologically reasonable dynamic strategy, helping to effectively prevent overfitting and underfitting\cite{zhao2022toward,zhao2021dynamically,zhao2017towards}. Such methods are only suitable for shallow artificial neural networks (ANNs), and the pure biological brain development mechanism has not been well understood and referenced.

Spiking Neural Networks (SNNs) are considered to be the third generation neural networks\cite{Maass1997Networks}, with spike event-driven computation, spatio-temporal joint information processing and high biological plausibility\cite{gerstner2002spiking}, which is more in line with the processing mechanism of the brain nervous network. Therefore, learning from the adaptive pruning mechanism of brain development is an effective way to prune SNNs, which is also lacking in current studies. Many existing methods simply apply pruning methods in ANNs to SNNs\cite{chen2021pruning,deng2021comprehensive,chen2023resource,meng2023efficient,chakraborty2024sparse}, which ignore the unique information processing with binary spikes of SNNs, thus limiting the performance of pruned SNNs. Some more biological SNNs pruning methods use spike-timing dependent plasticity (STDP) as an evaluation criterion, dynamically prune synapses with smaller weights or decayed weights in shallow SNNs \cite{rathi2018stdp,shi2019soft,nguyen2021connection,liu2022dynsnn}. Essentially, STDP as a local unsupervised plasticity mechanism is hard to be applied to deep SNN learning and far from the brain's pruning mechanism. Besides, these methods are limited to spatial pruning, ignoring the SNN's unique temporal dimension.

Although these attempts have become a feasible way of compressing deep networks, they draw little inspiration from the brain's development. Substantial efforts are still needed toward studying developmental plasticity-inspired deep networks, as only such biologically interpretable and plausible methods have the potential to approach the highly efficient brain nervous system. This paper aims to incorporate multi-scale spatio-temporal developmental plasticity mechanisms into both DNNs and DSNNs and answer how much the brain's adaptive pruning mechanism helps to better shape the network's structure. 

As the brain ongoing learns, neurons dynamically stretch out multiple dendrites for receiving information, and some dendritic spines form synapses through specific connections\cite{gray1959axo}. Synaptic plasticity between pre-synaptic and post-synaptic neurons contributes to the connectivity and efficiency of neural circuits that support learning, memory, and other cognitive abilities\cite{hebb1949first,skaliora2002experience}. During developmental pruning of the brain, dendritic spines, synapses, and neurons are continuously strengthened or decayed or even death according to the ``use it or lose it, gradually decay" principle\cite{bruer1999neural}. Dendritic spines formation (or enlargement) and elimination (or shrinkage) depend on the temporal spiking activity of the post-synaptic neuron: repeated inductions of long-term potentiation (LTP) lead to dendritic spines formation (or enlargement), whereas long-term depression (LTD) coupled with the elimination (or shrinkage) of spines\cite{toni1999ltp,becker2008ltd}. Repeated activation and frequent use after LTP also strengthen neuronal activity levels and synaptic efficacy. All these temporal event-driven developmental plasticity mechanisms induce adaptive pruning according to synaptic and neuronal efficacy. Specifically, brain pruning principles include: 1) Synapses and neurons that are rarely used are more likely to be eliminated during the pruning process\cite{chugani1996neuroimaging}. 2) Unimportant and redundant synapses and neurons are first gradually decayed and eventually pruned away\cite{colman1997alterations}. 3) Dendritic spine elimination precedes synaptic pruning, and synaptic pruning precedes neural death\cite{trachtenberg2002long,furber1987naturally}.

In this paper, we propose a generalized adaptive pruning algorithm for SNNs and DNNs inspired by the developmental plasticity of the brain. The proposed algorithm integrates multi-scale spatio-temporal developmental plasticity as the importance measure, and combines the pruning strategy of ``use it or lose it, gradually decay" to dynamically eliminate redundancy progressively evolving brain-inspired compact neural circuits and network architectures. The main contributions of this paper are summarized as follows:

\begin{enumerate}
	\item[$\bullet$]Synthesizing dendritic spine dynamic plasticity, local Bienenstock-Cooper-Munros (BCM)  \mbox{\cite{bienenstock1982theory}} synaptic plasticity, and activity-dependent neural spiking trace, we propose a generalized adaptive pruning algorithm to dynamically prune inactive synapses and neurons in both SNNs and DNNs.
 
	\item[$\bullet$] For event-driven SNNs, we spatially prune neurons and synapses based on temporal spiking traces. Inspired by the temporal attention mechanism of the brain, we further implement temporal dimension pruning for SNNs to reduce energy consumption even more.
 
	\item[$\bullet$] We introduce the proposed algorithm into both DNNs and SNNs improving convergence speed and accuracy while reducing energy consumption for various temporal and spatial datasets. Specifically, the proposed  spatio-temporal pruning method brings state-of-the-art (SOTA) performance with only 6.00\% energy consumption for SNNs on temporal datasets.

\end{enumerate}

The remainder of this paper is organized as follows. In Section \mbox{\ref{sec:meth}}, we present the proposed DPAP framework in detail. In Sections \mbox{\ref{expe}}, we evaluate the performance of DPAP on SNNs and DNNs. We conduct a series of discussions and analyses in Section \mbox{\ref{diss}}. Finally, we conclude our findings in Section \mbox{\ref{conc}}.

\section{Developmental Plasticity-inspired Adaptive Pruning Algorithm}
\label{sec:meth}
In this section, we present the proposed developmental plasticity-inspired adaptive pruning algorithm generalized for SNNs and DNNs, as shown in Fig. \mbox{\ref{method fig}}. We describe the overall framework of the proposed learning-while-pruning algorithm. Then, we provide computational details of the bioplasticity-based importance. Finally, we introduce the proposed adaptive pruning strategy in SNNs and DNNs respectively.

\subsection{The overall learning-while-pruning framework}

DPAP adaptively prunes irrelevant synapses and neurons
during the ongoing learning process without pre-training and re-training. According to the “use it or lose it”
principle, the DPAP takes inspiration from multiscale brain
pruning mechanisms, including local synaptic plasticity,
activity-dependent neural spiking trace, and dendritic spine
dynamic plasticity:


\begin{figure*}[t]
	\centering 
	\includegraphics[width=1\linewidth]{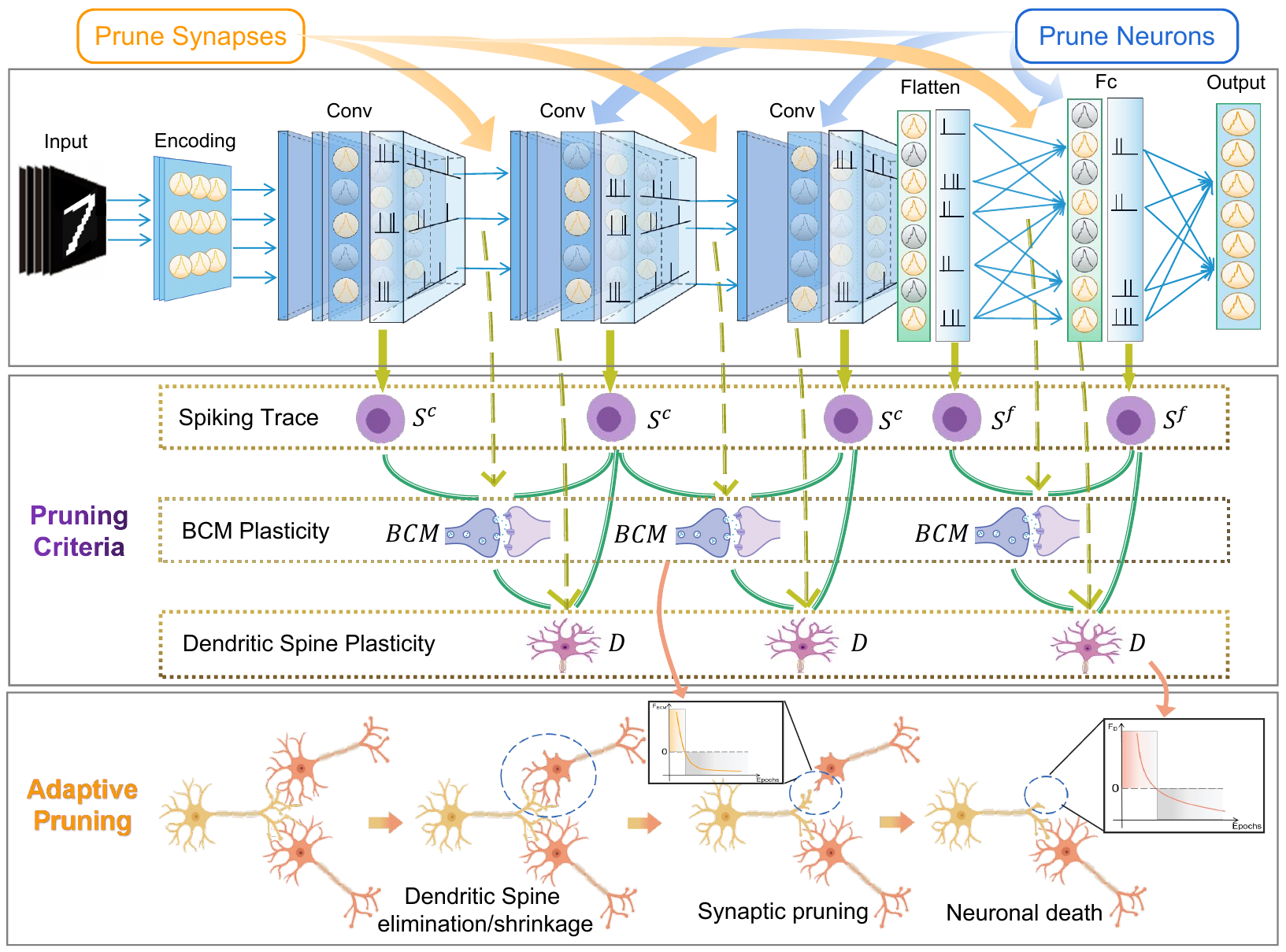}
	\caption{\textbf{The procedure of DPAP method.} The SNN structure \textbf{(top block)} consists of convolutional layers and fully connected layers. Pruning criteria \textbf{(middle block)} contains trace-based BCM plasticity for synapses and dendritic spine plasticity for neurons. Adaptive pruning \textbf{(bottom block)} gradually prunes decayed synapses and neurons according to survival function. The orange graph represents the survival function of the pruned synapse, and the red graph represents the survival function of pruned neurons.}
	\label{method fig}
\end{figure*}

\subsubsection{Trace-based BCM synaptic plasticity} We employed trace-based BCM synaptic plasticity~\cite{bienenstock1982theory} to measure the importance (or efficacy) of synapses, as BCM can induce LTP or LTD based on the activity of pre-synaptic and post-synaptic neurons, which is consistent with the criterion of brain synaptic pruning. Furthermore, the introduction of spiking trace makes the later timestep the neural firing, the stronger the correlation, which temporal information distinguishes the magnitude of synaptic or neuronal activity. Trace-based BCM is not only determined by the current spiking traces of pre-synaptic and post-synaptic neurons, but also takes into account the average of the historical activity of the post-synaptic neuron. Thus, the synaptic importance increases when the trace of the pre-synaptic neuron is sufficient to activate the post-synaptic neurons’ trace above the historical threshold. Otherwise, the synaptic importance declines.

\subsubsection{Activity-dependent Neural Spiking Trace} For the unique spatio-temporal information fusion of SNNs, the activity level of the neuron is measured by the spiking trace, which takes into account the spiking sequence in the previous period and the firing state at the current moment. Pre- and post-synaptic spiking traces model the effects of  N-methyl-D-aspartate (NMDA) and $Ca^{2+}$ on biological neuronal activity, respectively\mbox{~\cite{pfister2006triplets,yamazaki2022spiking}}. At any time, the neural spiking trace $S$ is accumulated by 1 when the neuron emits a spike, otherwise, the trace gradually decays with a time constant $\tau$. In DNNs, spiking traces are represented by neuron activation outputs.

\subsubsection{Dendritic Spine Dynamic Plasticity} Neurons extend a large number of dendritic spines to receive pre-synaptic information, and the density and volume of dendritic spines could reflect the activity level of neurons. For dendritic spines, some spines form synapses with pre-synaptic axon terminals to receive pre-synaptic spike signal transmission, and other spines are also expanding and shrinking in preparation for the formation of synaptic connections. Thus, the dynamic plasticity of dendritic spines incorporates the BCM synaptic plasticity with the neural spiking traces. Dendritic spines provide a comprehensive measure of the efficacy and importance of the neurons.

Next, the proposed adaptive pruning strategy is fed with synaptic importance measured by BCM synaptic plasticity and neuronal importance measured by dendritic spine plasticity to dynamically prune neurons and synapses. Drawing on the “gradually decay or even die” mechanism in brain developmental pruning, we define a survival function to decide whether synapses or neurons are removed. The survival function considers continuous changes in the importance of neurons and synapses, and only neurons and synapses with negative values of the survival function are permanently deleted. Thus, DPAP could ensure that the pruned synapses and neurons are redundant and unimportant, and naturally more biologically plausible.

For an epoch training of SNN and DNN, we follow the common direct training algorithm for SNNs and the backpropagation algorithm for DNNs to update weight gradient in the learning process of each batch. Only after all batch is completed, we calculate the importance of neurons and synapses and prune the network redundancy according to the proposed adaptive pruning strategy. For convolutional layers, we adopt structured pruning that treats each channel as an overall neuron population to prune. The DPAP method computational details are in the following sections.

\subsection{Biologically Plausible Pruning Criteria for SNNs}
In SNN, we used the common leaky integrate-and-fire (LIF)  spiking neurons\mbox{~\cite{Lapicque1999L}} as the basic unit and trained the network directly with the surrogate gradient algorithm\mbox{~\cite{wu2018spatio,zeng2022braincog}}. Event-driven temporal characterization is the unique advantage making SNN bio-interpretable and energy-efficient. Hence, both our synaptic and neuronal pruning criteria in SNN are calculated based on the temporal spiking sequences.
\subsubsection{Activity-dependent neural spiking trace} To synthesize the spike firing of the neuron at all timesteps, we introduced the spiking trace to represent the neuronal activity level. At the t-th timestep, if the neuron fires, its spiking trace $S$ will be added by 1. Otherwise, its spiking trace decayed with time constant $\tau=0.5$. For the fully connected layer, the spiking trace of the neuron $i$ in layer $f$ at timestep $t+1$ is calculated:

\begin{equation}
	\label{t}
	S_{i}^{t+1,f}=\tau S_{i}^{t,f}+o_{i}^{t+1,f}
\end{equation}
Considering the structural characteristics of the convolutional layer, we regard each channel as an entire population of neurons. The neural spiking trace for the convolution layer $c$ which has $C\times N\times N$ neurons is calculated as follow:

\begin{equation}
	\label{tc}
	S_{i}^{t+1,c}=\tau S_{i}^{t,c}+\sum^{N,N}_{k=1}o_{i}^{t+1,c}
\end{equation}
Therefore, the more spikes a neuron fires in its given timesteps, the larger the spiking trace.

\subsubsection{Trace-based BCM synaptic plasticity} The trace-based BCM not only calculates the spiking traces in the current epoch, but also considers the historical neuronal traces. The synaptic importance is jointly measured by the pre-synaptic and post-synaptic spiking traces. In particular, post-synaptic neurons are considered more important, the difference between their spiking trace and the sliding threshold $\theta$ determines the direction of the importance update. The trace-based BCM of batch $b$ is calculated by:

\begin{equation}
	\label{bcm}
	BCM_{pre-post}^{b}=S_{pre}^{T}\cdot S_{post}^{T}\cdot\left (S_{post}^{T}- \theta \right )
\end{equation}
where $S_{pre}^{T}$ and $S_{post}^{T}$ are the pre- and post-synaptic spiking trace over $T$ timesteps, respectively. The sliding threshold $\theta$ is the historical average of post-synaptic neuronal activity, as shown in Eq \ref{theta}. It determines the direction of LTP and LTD synaptic plasticity.

\begin{equation}
	\label{theta}
	\theta=\frac{ \theta *\left ( Num-1 \right )+ S_{post}}{Num}
\end{equation}
where $Num$ is the number of all the batches experienced from the beginning of learning. At the batch $b$ in the epoch $e$, $Num$ is the following value:

\begin{equation}
	\label{NUM}
	Num=e*N_{batch}+b
\end{equation}
where $N_{batch}$ is the number of the batch in an epoch. For each epoch, the synaptic importance pruning criteria $BCM^{e}$ is calculated as the sum of all batches:

\begin{equation}
	\label{bcme}
	BCM^{e}=\sum^{N_{batch}}_{b=1} BCM_{pre-post}^{b}
\end{equation}

\subsubsection{Dendritic spine dynamic plasticity} Dendritic spine plasticity is jointly determined by the neural spiking traces and the trace-based BCM synaptic plasticity. Therefore, for each epoch,  dendritic spine dynamic plasticity $D^{e}$ as the the neuronal importance pruning criteria is calculated as follow:

\begin{equation}
	\label{d}
 \begin{aligned}
	D^{e}&=\sum^{N_{batch}}_{b=1}S_{post}^{T} * \sum^{N_{pre}}_{j=1} BCM^{e}\\
 &=\sum^{N_{batch}}_{b=1}(S_{post}^{T} \sum^{N_{pre}}_{j=1} S_{pre}^{T}\cdot S_{post}^{T}\cdot\left (S_{post}^{T}- \theta \right ))
 \end{aligned}
\end{equation}
where ${N_{pre}}$ is the number of pre-synaptic channels or neurons. In summary, both synaptic and neuronal importance are calculated relying on the neuronal spiking traces, that is, the spiking information in the temporal dimension unique to the SNN.

\subsubsection{Timestep importance ciriteria} When processing temporal sequential data, the brain adapts to accurately capture valid events in key times\mbox{~\cite{raymond1992temporary,coull2004fmri}}. Spatio-temporal integrated SNN excels at processing event-driven neuromorphic temporal datasets, but the cumulative computation of multiple timesteps greatly increases the energy consumption of SNN. Therefore, inspired by the temporal attention mechanism of the brain, we design SNN-unique temporal dimension pruning to eliminate redundant timesteps. 

Specifically, the importance of the current timestep depends on the neuronal activity at that timestep. It is determined by the acceptable spikes already fired by the pre-synaptic neurons before timestep $t$ and spike from this neuron at the current timestep. We follow the proposed spike-based importance, and the timestep $t$ importance of neuron $i$ is calculated by the spiking trace of the pre-synaptic neuron during the $0 \sim t$ timesteps and the spike of the post-synaptic neuron at the timestep $t$, as follow: 

\begin{equation}
	\label{time}
	Time^{t}=\sum^{N_{pre}}_{j=1}S_{pre}^{0 \sim t}\cdot O_{post}^{t}
\end{equation}

\subsection{Developmental Plasticity-inspired Adaptive Pruning}
\subsubsection{DPAP in SNNs} The DPAP method prunes unimportant synapses, neurons and timesteps according to their importance $BCM^{e}$, $D^{e}$ and $Time^{t}$. Inspired by the pruning mechanism of brain development, we designed the survival function for neuron $F_{D}$, synapse $F_{BCM}$ and timestep $F_{Time}$. At the beginning of learning, we initialized the survival functions as the constant $\beta$. Take synapses for example, the trace-based BCM synaptic plasticity linearly normalized in each epoch as follow:

\begin{equation}
	\label{db}
	\delta_{BCM}^{e}=2*Normalized(BCM^{e})- \epsilon
\end{equation}
where $\epsilon$ is the decay value. Then, we calculated the update value of synapse survival function $\Delta F_{BCM}^{e}$:

\begin{equation}
	\label{dbb}
	\Delta F_{BCM}^{e}=\left\{\begin{matrix}
		\delta_{BCM}^{e}+C, & \delta_{BCM}^{e} \geq 0 \\ 
		\delta_{BCM}^{e}, & otherwise
	\end{matrix}\right.
\end{equation}
For the reliability of pruning, we protected the synapses with positive $\delta_{BCM}^{e}$ by extra increasing by constant $C$ ($C=5$ in convolutional layers, $C=2$ in fully connected layers). During the whole learning process, the survival function $F_{BCM}$ are updated as the decay rate $\eta$:
\begin{equation}
	\label{f}
	F_{BCM}=\gamma F_{BCM}+e^{-\frac{epoch}{\eta}}\Delta F_{BCM}^{e}  
\end{equation}
where $\gamma=0.999$ is the decay constant of the survival function. 

The neuron survival function $F_{BCM}$ and the timestep survival function  $F_{Time}$ are calculated and updated consistent with the above synapse survival function $F_{BCM}$. Finally, we prune the synapses whose $F_{BCM}<0$ by setting their weight $w_{ij}=0$, prune the neurons whose $F_{D}<0$ by setting all their pre-synaptic weight $W_{i}=0$, and prune the neuronal timestep whose $F_{Time}<0$ by forcing the neuron to no longer firing spike at this timestep. 

\vspace*{1\baselineskip} 

\subsubsection{DPAP in DNNs} For traditional DNNs, we employed the RELU activation function and the cross-entropy loss function. Since DNNs do not have temporal dimension information which is distinct from SNNs, we only prune synapses and neurons in the spatial dimension. Besides, the neural spiking trace of the neuron in DNN is defined as the output of the neuron after activation function. For the fully connected layer:

\begin{equation}
	\label{tdf}
	S_{i}^{DNN,f}=RELU \left (W_{f}x_{j}^{f-1}+b_{f} \right)
\end{equation}
For the convolution layer with $C\times N\times N$ neurons:

\begin{equation}
	\label{tdc}
	S_{i}^{DNN,c}=\sum^{N,N}_{k=1} RELU \left (W_{c} \otimes x_{j}^{c-1}+b_{c} \right)
\end{equation}
where $x_{j}^{f-1}$ and   $x_{j}^{c-1}$  is the inputs of the layer.

For each batch, the  trace-based BCM synaptic plasticity $BCM_{pre-post}^{b}$ is calculated same as Eq \ref{bcm} based on  pre- and post-synaptic spiking trace as Eq \ref{tdf},\ref{tdc}. The sliding threshold is update as Eq \ref{theta}. For each epoch, the $BCM^{e}$ is the sum of $BCM_{pre-post}^{b}$ and the dendritic spine $D^{e}$ is calculated same as Eq \ref{d}. Besides, the adaptive pruning strategy in DNNs is the same as in SNNs as Eq \ref{db} to \ref{f}.

\vspace*{1\baselineskip} 

For a feed-forward SNN or DNN, we provide the  generic detailed procedure for the DPAP algorithm as Algorithm \ref{alg1}:

\begin{algorithm}[t]
	\caption{The DPAP Algorithm}
	\label{alg1}
	\KwIn{Base redundant model; Training set $Y$.}
	\KwOut{The pruned model.}
	Initialize: randomly initialize network weight $W$, initialize $F_{BCM}=\beta$, $F_{D}=\beta$ for each layer.\\
	\For{$e=0$; $e<Epoch$;$e++$}{
		\For{$b=0$; $b<N_{batch}$;$b++$}{
        //Learning network weights\\
		Forward propagation getting outputs;\\
  	Backward propagation by surrogate gradient for SNN  or standard BP for DNN;\\
        //Collecting variables needed for pruning\\
	Calculate spiking trace as Eq \ref{t},\ref{tc} or Eq\ref{tdf},\ref{tdc};\\
	Calculate $BCM_{pre-post}^{b}$  as Eq \ref{bcm},\ref{theta};\\
		}
  //Pruning network redundancy\\
	Calculate $BCM^{e}$  and  $D^{e}$ as Eq \ref{bcme} and \ref{d};\\
	Calculate the $F_{BCM} $ and $ F_{D}$  as Eq \ref{db} to Eq \ref{f};\\
	Prune neurons with $ F_{D}<0$ and synapses with $F_{BCM}<0$;\\
	}
\end{algorithm}

\begin{figure*}[t] 
	\centering  
	\includegraphics[width=1\linewidth]{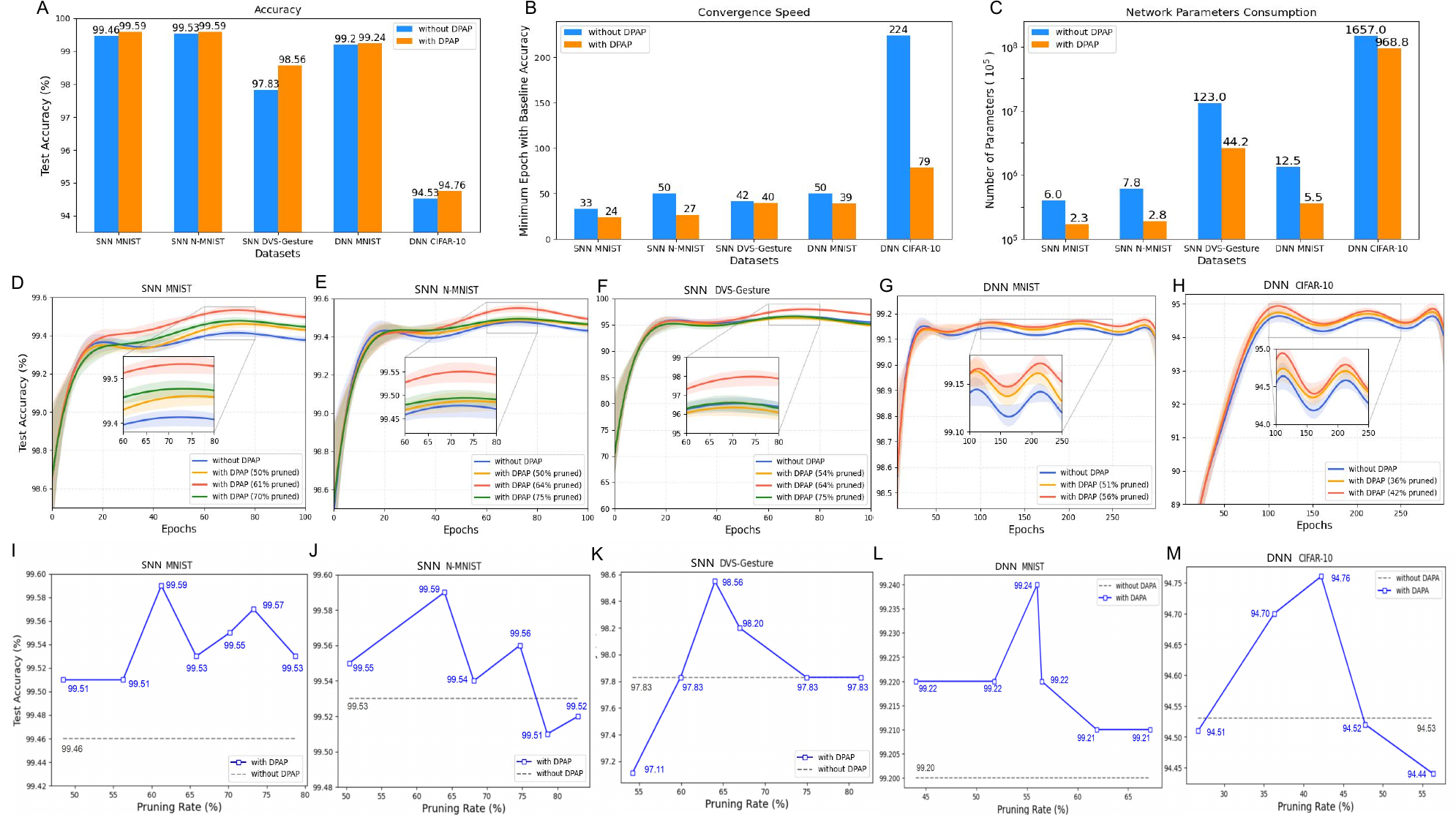} 
	\caption{\textbf{The effectiveness of introducing DPAP to DSNNs and DNNs.} \textbf{(A)} to \textbf{(C)}: The test accuracy, convergence speed and energy consumption achieved with and without DPAP, respectively. \textbf{(D)} to \textbf{(H)}: Under different pruning rates, the accuracy changes with the iteration process for different datasets. \textbf{(I)} to \textbf{(M)}: The test accuracy achieved by DPAP with different pruning rates for different datasets.}
	\label{snn fig}
\end{figure*}
\begin{table}[!h]
  \centering
  \caption{Initial structures of different tasks}
  \setlength{\tabcolsep}{0.95mm}{
    \begin{tabular}{cll}
    \toprule
    \toprule
      & \textbf{Dataset} & \multicolumn{1}{c}{\textbf{Initial structure}} \\
    \midrule
    \multirow{5}[2]{*}{SNN} & N-MNIST & 15C3-AvgPool2-40C3-AvgPool2-300FC-10FC \\
      & DVS-Gesture & 15C3-AvgPool2-40C3-AvgPool2-300FC-10FC \\
      & MNIST &15C3-AvgPool2-40C3-AvgPool2-300FC-10FC \\
      & CIFAR-10 & \makecell{128C3-BN-128C3-BN-MaxPool2-256C3-BN-256C3\\-BN-MaxPool2-512C3-BN-512C3-BN-512FC-10FC} \\
      &ImageNet & ResNet-18 \\
    \midrule
    \multirow{3}[2]{*}{DNN} & MNIST & 20C5-MaxPool2-50C5-MaxPool2-500FC-10FC \\
      & CIFAR-10 & VGG16 \\
      & ImageNet & ResNet-50 \\
    \bottomrule
    \bottomrule
    \end{tabular}}
  \label{str}%
\end{table}%

\section{EXPERIMENTS}
\label{expe}
In this section, we performed extensive experiments and comparisons with others validating that the DPAP method could remarkably reduce energy consumption while improving convergence speed and accuracy for both DSNNs and DNNs\footnote{Our code is available at: https://github.com/BrainCog-X/Brain-Cog/tree/main/examples/Structural\_Development/DPAP}.
\subsection{Experimental settings}
We introduce spatial pruning and temporal pruning in the temporal neuromorphic datasets (N-MNIST\mbox{\cite{orchard2015converting}}, DVS-Gesture\mbox{\cite{amir2017low}})  for SNNs. The N-MNIST Dataset is captured by the neuromorphic vision sensor from original MNIST, and the DVS-Gesture dataset has 11 different gestures of 29 subjects captured by the DVS camera, with 1176 training samples and 280 testing samples. Moreover, we validate spatial pruning in the spatial datasets  (MNIST\mbox{\cite{lecun1998mnist}}, CIFAR-10 \mbox{\cite{krizhevsky2009learning}},ImageNet\mbox{\cite{deng2009imagenet}}) for SNNs and DNNs. The initial network structures used for the above tasks are shown in Table \mbox{\ref{str}}. 

The evaluation criteria of convergence speed is defined by the minimum epochs required to reach the highest accuracy of the initial baseline network. As in \mbox{\cite{chakraborty2021fully}}, the energy consumption of SNN is defined as: 
\begin{equation}
	\label{es}
	E_{SNN}=FLOPS_{SNN}*E_{AC}*T
\end{equation}
where $E_{AC}$ is the energy consumption of accumulate (AC) operations and $FLOPS_{SNN}$ is the floating point operations per second. The energy consumption of SNN after pruning is:
\begin{equation}
	\label{es2}
	E_{SNN}^{p}=(1-\rho_w)*FLOPS_{SNN}*E_{AC}*(1-\rho_t)*T
\end{equation}
where $\rho_w$ and $\rho_t$ are the weight sparsity and timestep sparsity respectively. For DNNs, the spatial weights sparsity represents the energy reduction. 

\begin{table*}[!h]
	\footnotesize
	\centering
	\caption{Performance Comparison for SNN on temporal datasets N-MNIST and DVS-Gesture.} 
	\setlength{\tabcolsep}{0.8mm}{
		\begin{tabular}{ccccccccc}
			
			\toprule
			\toprule
			\textbf{Dataset} & \textbf{Method} & \textbf{Structure} & \textbf{Pruning Methods}& \makecell{\textbf{Weight} \\\textbf{sparsity}} & \makecell{\textbf{Timestep}\\ \textbf{sparsity}} &\makecell{\textbf{Energy}\\ \textbf{Consumption}} & \textbf{Accuracy} & \makecell{\textbf{Accuracy}\\ \textbf{Loss}} \\
			\midrule
			
			\multirow{9}[6]{*}{N-MNIST} & \multirow{2}[2]{*}{ADMM-based~\cite{deng2021comprehensive}} & \multirow{2}[2]{*}{LeNet-5} & \multirow{2}[2]{*}{Spatial pruning} & 50.00\% &0.00\% & 50.00\%& 98.34\%  & -0.61\%  \\
			&   &   &   & 75.00\% &0.00\% & 25.00\% & 96.83\%  & -2.12\%  \\
			\cmidrule{2-9}      & \multirow{2}[2]{*}{Grad R~\cite{chen2021pruning}} & \multirow{2}[2]{*}{2 Conv 2 FC} & \multirow{2}[2]{*}{Spatial pruning} & 65.00\% &0.00\% & 35.00\% & 99.37\%  & 0.54\%  \\
			&   &   &   & 75.00\% &0.00\% & 25.00\% & 98.56\%  & -0.27\%  \\
			\cmidrule{2-9}      & \multirow{5}[2]{*}{\textbf{Our DPAP}} & \multirow{5}[2]{*}{2 Conv 2 FC} & \multirow{3}[2]{*}{Spatial pruning} & 50.41\% &0.00\% & 49.59\% & 99.55\%  & 0.02\%  \\
			&   &   &   & 63.95\% &0.00\% & 36.05\% & 99.59\%  & 0.06\%  \\
			&   &   &   & 74.66\% &0.00\% & 25.34\% & 99.56\%  & 0.03\%  \\
            \cmidrule{4-9}
            &   &   &   \multirow{2}[2]{*}{Spatial-temporal pruning} & 62.01\%  &  53.39\% &  17.70\% & 99.53\% &0.00\%\\
			&   &   &   & 70.63\%  &  62.25\% &  11.09\% & 99.55\% &0.02\%\\
			\midrule
			\multirow{9}[6]{*}{DVS-Gesture} & \multirow{2}[2]{*}{Deep R~\cite{chen2021pruning}} & \multirow{2}[2]{*}{2 Conv 2 FC} & \multirow{2}[2]{*}{Spatial pruning} & 50.00\% &  0.00\%  & 50.00\% & 81.59\%  & -2.53\%  \\
			&   &   &   & 75.00\% &  0.00\% & 25.00\%& 81.23\%  & -2.89\%  \\
			\cmidrule{2-9}      & \multirow{2}[2]{*}{Grad R~\cite{chen2021pruning}} & \multirow{2}[2]{*}{2 Conv 2 FC} & \multirow{2}[2]{*}{Spatial pruning} & 50.00\% &  0.00\% & 50.00\% & 84.12\%  & 0.00\%  \\
			&   &   &   & 75.00\% &  0.00\% & 25.00\% & 91.95\%  & 7.83\%  \\
			\cmidrule{2-9}      & \multirow{5}[2]{*}{\textbf{Our DPAP}} & \multirow{5}[2]{*}{2 Conv 2 FC} & \multirow{3}[2]{*}{Spatial pruning} & 64.03\% &  0.00\% &35.97\% & 98.56\%  & 0.73\%  \\
			&   &   &   & 66.98\% &  0.00\%  & 33.02\%& 98.20\%  & 0.37\%  \\
			&   &   &   & 81.45\% &  0.00\%  & 18.55\% & 97.83\%  & 0.00\%  \\
               \cmidrule{4-9}

           &   &   &    \multirow{2}[2]{*}{Spatial-temporal pruning} & 57.67\%  &  49.93\% & 21.19\% & 98.20\% &0.37\%\\
			&   &   &   & 81.72\%  &  67.20\% & 6.00\%& 98.20\% &0.37\%\\

			\bottomrule
			\bottomrule
	\end{tabular}}
	\label{t1}%
\end{table*}%

\subsection{DPAP reduces energy consumption, improves the performance, and speeds up the convergence rate of DSNNs}

We first tested the effects of introducing DPAP on the accuracy (Fig.\ref{snn fig} A), convergence speed (Fig.\ref{snn fig} B) and energy consumption (Fig.\ref{snn fig} C) of SNNs for three benchmark datasets: MNIST, N-MNIST, DVS-Gesture. From Fig.\ref{snn fig} A and Fig.\ref{snn fig} C, we found that introducing DPAP could slightly improve the accuracy (averaged by $\sim$0.31\%) compared to the initial network without DPAP, while the networks are extremely compressed (averaged by $\sim$63\%). Compared to the initial network without DPAP, the pruned network with DPAP helped to elevate the test accuracy from 99.46\% to 99.59\% with only 38.75\% energy consumption for MNIST, and from 99.53\% to 99.59\% with 36.05\% energy consumption for N-MNIST, and from 97.83\% to 98.56\% with 35.97\% energy consumption for DVS-Gesture, respectively. These results also highlighted the superiority of our DPAP method on temporal neuromorphic datasets (such as DVS-Gesture). Besides, the convergence speeds of DSNNs using DPAP are significantly faster  (speeds up by $\sim$1.4$\times$ as shown in Fig.\ref{snn fig} B) than that without DPAP. Especially on N-MNIST dataset, DPAP achieves more outstanding advantages in accelerating convergence (speeds up by $\sim$1.85$\times$). In summary, DPAP could elevate the efficiency of DSNNs by extremely compressing the network (up to 64.03\%) and speed up learning (up to 1.85$\times$) with even relative accuracy improvement (up to 0.73\%).

Furthermore, we compared the test accuracy of different datasets during learning by DPAP under different pruning rates and without DPAP (Fig.\ref{snn fig} D-F after polynomial fit). Here, the different pruning rates are affected by two parameters (decay value $\epsilon$ and decay rate $\eta$) of the survival function in the DPAP method, where the faster neurons and synapses decay, the greater the pruning rate of the final network. Similar conclusions can be obtained with different datasets and pruning rates, that is, DPAP starts to make sense between 20-40 epochs, and then gradually widens the performance gap with the network without DPAP, and eventually achieves the highest performance between 60-80 epochs. Moreover, DPAP could achieve comparable or even better performance under different pruning rates, especially for the MNIST and N-MNIST datasets, which show more adaptability and stability to different pruning rates. 

\begin{table}[!h]
	\footnotesize
	\centering
	\caption{Performance Comparison for SNN on spatial datasets. }
 \begin{threeparttable}
	\setlength{\tabcolsep}{0.6mm}{
		\begin{tabular}{clcccc}
			
			\toprule
			\toprule
			\textbf{Dataset} & \textbf{Method} & \textbf{Structure} &  \textbf{Sparsity} & \textbf{Accuracy} & \textbf{AccLoss} \\
			\midrule
			\multirow{18}[12]{*}{MNIST} & Online APTN\tnote{\dag} ~\cite{guo2020unsupervised} & 2 FC & 90.00\%  & 86.53\%  & -3.87\%  \\
			& Threshold based\tnote{\dag}~\cite{shi2019soft}  &  1 FC & 70.00\%  & 75.00\%  & -19.05\%  \\
			& Threshold based\tnote{\dag}~\cite{rathi2018stdp}  & 2 FC & 92.00\%  & 91.50\%  & -1.70\%  \\
			& Threshold based \tnote{\ddag}~\cite{neftci2016stochastic} &  2 FC & 74.00\%  & 95.00\%  & -0.60\% \\
			\cmidrule{2-6}      & \multirow{2}[2]{*}{ADMM based~\cite{deng2021comprehensive}}  & \multirow{2}[2]{*}{LeNet-5 } & 50.00\%  & 99.10\%  & 0.03\%  \\
			&   &   &    75.00\%  & 96.84\% & -2.23\% \\
			\cmidrule{2-6}      & \multirow{2}[2]{*}{Deep R~\cite{chen2021pruning}}  & \multirow{2}[2]{*}{2 FC} & 62.86\%  & 98.56\%  & -0.36\%  \\
			&   &   &    86.70\% & 98.36\% & -0.56\% \\
			\cmidrule{2-6}      & \multirow{2}[2]{*}{Grad R~\cite{chen2021pruning}}  & \multirow{2}[2]{*}{2 FC} & 74.29\% & 98.59\% & -0.33\% \\
			&   &      & 82.06\% & 98.49\% & -0.43\% \\
			\cmidrule{2-6}      & DynSNN~\cite{liu2022dynsnn}  &{3 FC} & 57.40\% & 99.23\% & -0.02\% \\
			\cmidrule{2-6}      & DynSNN\tnote{*}~\cite{liu2022dynsnn}  & LeNet-5  & 61.50\% & 99.15\%  & -0.35\% \\
			\cmidrule{2-6}      & \textbf{Our DPAP}  & 2 FC & 77.36\%  & 98.74\% & -0.07\%  \\
			\cmidrule{2-6}      & \multirow{2}[2]{*}{\textbf{Our DPAP}}  & \multirow{2}[2]{*}{2Conv2FC} 
                & 61.25\%  & 99.59\%  & 0.13\% \\
			&   &    & 84.22\% & 99.56\% & 0.10\% \\
			\midrule
			\multirow{8}[8]{*}{CIFAR-10} & \multirow{2}[2]{*}{ADMM based~\cite{deng2021comprehensive}} & \multirow{2}[2]{*}{7Conv2FC} & 40.00\% & 89.75\%  & 0.18\% \\
			&   &     & 60.00\%  & 88.35\%  & -1.18\%  \\
			\cmidrule{2-6}      & DynSNN\tnote{*}~\cite{liu2022dynsnn}  & ResNet-20 & 37.13\% & 91.13\% & -0.23\% \\
			\cmidrule{2-6}      & \multirow{2}[2]{*}{Grad R~\cite{chen2021pruning}} & \multirow{2}[2]{*}{6Conv2FC} & 71.59\% & 92.54\% & -0.30\% \\
			&   &      & 87.96\% & 92.50\% & -0.34\% \\
			\cmidrule{2-6}      & \multirow{2}[2]{*}{\textbf{Our DPAP}} & \multirow{2}[2]{*}{6Conv2FC} & 33.46\% & 94.27\%  & -0.27\% \\
			&   &      & 50.80\% & 93.83\% & -0.71\% \\
      \midrule
	\multirow{10}[10]{*}{ImageNet}&
 	\multirow{2}[2]{*}{ADMM based~\cite{deng2021comprehensive} }  & \multirow{2}[2]{*}{ResNet18} & 70.75\% & 59.48\% & -3.74\% \\
	&   &    &78.96\% & 55.85\% & -7.37\% \\
	\cmidrule{2-6}&
   \multirow{2}[2]{*}{Grad R~\cite{chen2021pruning}} & \multirow{2}[2]{*}{ResNet18} & 50.94\% & 60.05\% & -3.17\% \\
	&   &    &53.65\% & 24.62\% & -38.60\% \\
	\cmidrule{2-6}&
	\multirow{2}[2]{*}{\makecell{Unstructured\\Pruning}~\cite{shi2023towards}} & \multirow{2}[2]{*}{ResNet18} & 64.74\% & 61.89\% & -1.29\% \\
	&   &    &72.26\% & 60.00\% & -3.18\% \\
	\cmidrule{2-6}&
	\multirow{3}[2]{*}{\textbf{Our DPAP}} & \multirow{3}[2]{*}{ResNet18} & 22.69\% & 63.74\% & -2.00\% \\
	&   &    & 37.76\% & 63.35\% & -2.39\%\\
    &   &    & 51.71\% & 60.41\% & -5.33\%\\
			\bottomrule
			\bottomrule
	\end{tabular}}
         \begin{tablenotes}[para,flushleft]
            \item[] \textit{Training methods:} 
            \item[\dag] STDP \item[\ddag] Event-driven CD \item[*] ANN-to-SNN \item[] The rest use surrogate gradient.
        \end{tablenotes}
    \end{threeparttable}
	\label{t2}%
\end{table}%
Fig.\ref{snn fig} I-K illustrates the best performance achieved by DPAP with different pruning rates for different datasets. Unlike the conclusion found in most other works that the higher the pruning rate, the worse the performance, we observed that the accuracy peaks at about 60\% pruning ratio. Actually, at age two or three, a child’s brain has up to twice as many synapses as it will have in adulthood\cite{huttenlocher1979synaptic}. Synaptic pruning happens very quickly between ages 2 and 10\cite{huttenlocher1994synaptogenesis}. During this time, about 50 percent of the extra synapses are eliminated\cite{huttenlocher1990morphometric}. Subsequently, synaptic pruning continues through adolescence, but not as fast as before. The total number of synapses begins to stabilize\cite{huttenlocher1979synaptic}. In addition, “over-pruned” or “under-pruned” during brain development would lead to the occurrence of diseases, such as schizophrenia and autism spectrum disorders, respectively\cite{penzes2011dendritic,glausier2013dendritic,hutsler2010increased}. All these evidence reveals that our conclusions are more biologically reasonable, and closer to the pruning mechanism of brain development.
\begin{table}[t]
  \centering
  \caption{Performance comparison for DNN.}
  \setlength{\tabcolsep}{1.2mm}{
    \begin{tabular}{clcccc}
    \toprule
    \toprule
    \textbf{Dataset} & \textbf{Pruning Method} & \textbf{Structure} & \textbf{Sparsity} & \textbf{Accuracy} &\textbf{AccLoss} \\
    \midrule
    \multirow{4}[2]{*}{MNIST} 
          & BSP~\cite{zhao2021dynamically} & 2FC   & 83.38\% & 95.94\% & 0.33\%\\
      & NP~\cite{srinivas2015data} & LeNet    & 59.62\% & 98.98\% &-0.08\%\\
      &AL~\cite{srinivas2016learning}  & LeNet &  90.51\% & 99.04\% & -0.03\%\\
      & \textbf{Our DPAP} & 2Conv2FC  & 56.00\% & 99.24\%& 0.04\% \\ 
    \midrule
    \multirow{4}[2]{*}{CIFAR-10} 
      & AFP-F~\cite{ding2018auto}& VGG16  & 81.39\% & 92.87\% & -0.05\%\\
        &AAP~\cite{zhao2023automatic}  & VGG16 & 70.04\% & 93.37\% & -0.27\%\\
                & CP~\cite{he2017channel}  & VGG16 & 50.00\% & 93.67\% & -0.32\%\\
        & NS~\cite{liu2017learning}  & VGG16 & 51.00\% & 93.80\% & -0.19\%\\
        & ThiNet~\cite{luo2017thinet}  & VGG16 & 50.00\% & 93.85\% & -0.14\%\\
        & Li-pruned~\cite{li2016pruning}  & VGG16& 64.00\% & 94.40\% & 0.15\% \\
      & \textbf{Our DPAP}& VGG16 & 41.54\% & 94.76\% & 0.23\% \\
    \midrule
    \multirow{8}[2]{*}{ImageNet} 
     & Entropy~\cite{luo2017entropy} &ResNet50& 35.00\% & 70.84\% & -2.04\%\\
     & ThiNet~\cite{luo2017thinet}&ResNet50 & 56.00\% & 71.01\% & -1.87\%\\
        & GAL~\cite{lin2019towards}&ResNet50 & 16.90\% & 71.95\% & -4.20\%\\
& HRank~\cite{lin2020hrank} &ResNet50& 62.00\% & 71.98\% & -4.17\%\\
&ABCPruner~\cite{lin2020channel} &ResNet50& 68.00\% & 72.58\% & -3.42\% \\
  & SM~\cite{dettmers2020sparse}&ResNet50 & 82.00\% & 72.65\% & -1.36\%\\
 & PS~\cite{wang2020pruning}&ResNet50 & 78.60\% & 72.80\% & -4.40\%\\
 & {\textbf{Our DPAP}} &ResNet50 & 59.63\% & 73.26\% & -1.00\%\\
\bottomrule
			\bottomrule
    \end{tabular}%
    }
  \label{tab:addlabel}%
\end{table}%

\subsection{Comparison with existing state-of-the-art SNNs compression algorithms on five benchmark datasets}

Table. \mbox{\ref{t1}} and Table. \mbox{\ref{t2}} shows the comparison of the performance of different methods on temporal datasets and spatial datasets, respectively. For the temporal N-MNIST dataset, accuracy of spatial pruning drops when compressing the network by Grad R\mbox{\cite{chen2021pruning}} and ADMM-based\mbox{\cite{deng2021comprehensive}} pruning method, such as the accuracy reduces by 2.12\% when the network is compressed by 75.00\% for ADMM-based method\mbox{\cite{deng2021comprehensive}}. Our DPAP method achieves 74.66\% compression with even 0.03\% relative accuracy improvement, and reaches the maximum accuracy of 99.59\%  with 63.95\% pruning rate. When temporal pruning is added, we still have the accuracy up to 99.55\% under 70.63\% spatial weight sparsity and 62.25\% timestep sparsity. According to Eq. \mbox{\ref{es2}}, the energy required for the network after joint spatio-temporal pruning is only 11.09\% of the basic dense network.

For the DVS-Gesture dataset, the accuracy is very sensitive to Deep R\cite{chen2021pruning} pruning method, with 2.53\% accuracy drops at 50.00\% pruning rate, and with 2.89\% accuracy drops at 75.00\% pruning rate. Our DPAP could still improve the accuracy by 0.73\% while compressing the network up to 64.03\%. Moreover, even with only 18.55\% connections, the accuracy of our method is still up to 97.83\% (without any accuracy drop). With the similar spatial pruning rate of 81.72\% and temporal pruning rate of 67.20\%, the accuracy of joint spatio-temporal pruning is improved by 0.37\% compared to only spatial pruning reaching 98.20\%. Meanwhile, the network energy consumption is only 6.00\% of the basic dense network. These experimental results demonstrate that our proposed temporal pruning working jointly with spatial pruning further reduces the network energy consumption through suppressing the spiking firing without affecting the network performance. 

For the spatial MNIST dataset, other pruning methods compress the network at the cost of an accuracy drop. The ADMM-based pruning method\cite{deng2021comprehensive} loses 2.23\% accuracy when the network is compressed by 75.00\%. Our method could maintain slight performance improvement at different pruning rates ranging from 48.42\% to 84.22\%. Especially when the pruning rate reaches 84.22\%, our method can still achieve an accuracy improvement of 0.1\% (accuracy is up to 99.56\%). For the complex CIFAR-10 dataset, our DPAP method achieves 93.83\% accuracy at 50.80\%  pruning rate (with 0.71\% accuracy drop). Although there is a slight drop in accuracy, our method still achieves the highest accuracy of 94.27\% at the pruning rate of 33.46\%, outperforming the highest accuracy of 92.54\% with Gard R\cite{chen2021pruning} and the highest accuracy of 89.75\% with ADMM-based\cite{deng2021comprehensive} methods. For large-scale ImageNet dataset, our method achieves the outstanding accuracy of 63.35\% with 37.76\% pruning rate, which is higher than the next highest Unstructured Pruning\mbox{\cite{shi2023towards}} method of 1.46\%. Under the pruning rate of 51.71\%, DPAP still achieves a 60.41\% accuracy, which improves by 0.36\% compared to Grad R\mbox{\cite{chen2021pruning}} with similar pruning rate. 

To sum up, compared to other existing state-of-the-art SNNs compression algorithms, our pruning method shows obvious advantages, it can guarantee stable performance improvement under different pruning rates, and achieves SOTA effects on both performance and energy consumption for several benchmark datasets. 

\subsection{DPAP reduces energy consumption, improves the performance, and speeds up the convergence rate of DNNs}
To verify the generality of our developmental plasticity-inspired pruning model, we also examined the effects of introducing DPAP to DNNs on spatial datasets. Fig.\ref{snn fig} A-C also illustrates the accuracy, convergence speed and energy consumption of the network with DPAP and without DPAP for DNNs. Notably, DPAP could greatly compress the network (56.00\% compressed for MNIST, 41.54\% compressed for CIFAR-10) while relatively improving accuracy (0.04\% for MNIST, 0.23\% for CIFAR-10). In terms of convergence speed, for the MNIST dataset, the original network needs 50 epochs to reach the maximum accuracy of 99.20\%, while the network with DPAP can achieve the same accuracy at the 39th epoch. The learning speed accelerated by 1.28$\times$. Moreover, for the CIFAR-10 dataset, DPAP improves the convergence speed more than that without DPAP by 2.84$\times$. Furthermore, we compared the curve of test accuracy with different pruning rates achieved by DPAP or without DPAP and found that DPAP can consistently improve the performance of the network (Fig.\ref{snn fig} G,H after polynomial fit). Especially from 100-250 epochs, the accuracy of DNN with DPAP is significantly higher than the baseline DNN without DPAP.

\begin{figure*}[t] 
	\centering  
	\includegraphics[width=1\linewidth]{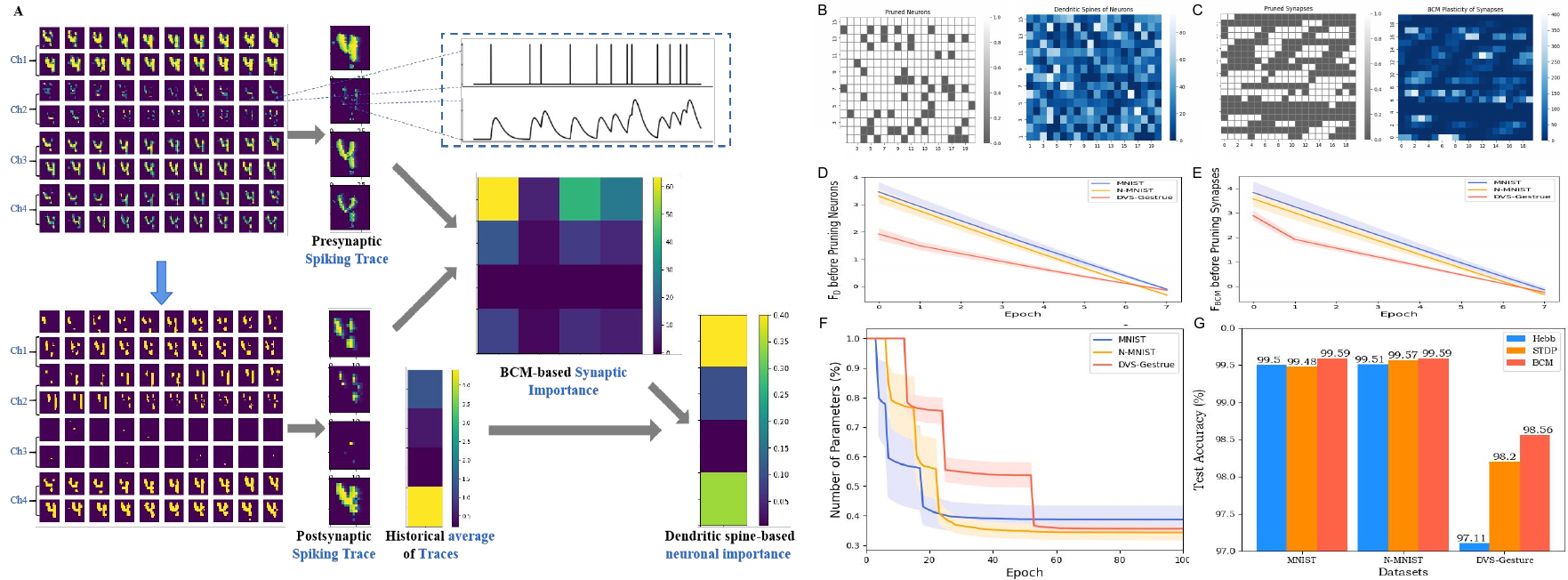} 
	\caption{\textbf{Analyses and visualization of the biological plausibility of DPAP.} \textbf{(A)}: Visualization of temporal spikes, spiking traces, and the importance of spatial pruning computed from temporal spiking traces. Take for example the first four channels of the first and second convolutional layers of the N-MNIST network with the timesteps of 20. \textbf{(B,C)}: The retained neurons and the average importance of all neurons (or synapses)  The white squares represent pruned neurons and synapses. \textbf{(D,E)}: During the 8 epochs before the neurons or synapses are pruned, the changing of survival function $F_{D}$  and $F_{BCM}$. \textbf{(F)}: The changing of network parameters during learning. \textbf{(G)}: Test accuracy comparison of different synaptic plasticity used.}
	\label{discuss}
\end{figure*}

With the datasets of MNIST and CIFAR-10, we also analyzed the accuracy of DNNs with DPAP under different pruning rates, as shown in Fig.\ref{snn fig} L,M. The results obtained a similar conclusion to DSNNs with DPAP, that as the pruning rate gradually increased, the accuracy showed a trend of increasing first and then decreasing, and formed a clear peak. Specifically, for the MNIST dataset, DNNs with DPAP reached the optimal balance of accuracy at 56.00\% pruning rate. Besides, the accuracy consistently exceeds baseline levels from 44.00\% to 67.17\% of the pruning rate. Even with 67.17\% pruning, DPAP can still achieve 99.21\% accuracy (improved by 0.01\%). For the CIFAR-10 dataset, DPAP can improve the accuracy of the network in the case of 36.31\% and 41.54\% compression, and achieves the highest performance of 94.76\% at 41.54\% pruning rate, while the performance is slightly lower (averaged by $\sim$0.04\%) than the baseline at 26.71\%, 47.78\%, and 56.37\% pruning rate (maintained an acceptable accuracy). As shown in Fig.\ref{snn fig} G,H, the introduction of DPAP to DNNs can achieve optimal effects when the pruning rate is about 48.77\%, which is consistent with the biological developmental mechanism. Furthermore, compared to other DNNs pruning algorithms under the same structure, our method achieves comparable performance. Particularly, for the large-scale dataset ImageNet, our DPAP method achieves an excellent 73.26\% accuracy (slightly lower than baseline DNN 1.00\%) at a biologically reasonable pruning rate of 59.63\%. At similar pruning rates, the DPAP accuracy is improved by 2.25\% and 1.28\% compared to ThiNet\mbox{\cite{luo2017thinet}} and HRank\mbox{\cite{lin2020hrank}}, respectively. Compared to the Pruning from scratch (PS)\mbox{\cite{wang2020pruning}}, which has the next highest accuracy, our DPAP improves by 0.46\%, while reducing the pruning accuracy loss by 3.40\%.  These results also illustrate the effectiveness of introducing the proposed brain development-inspired pruning approach (DPAP) to DNNs.

In conclusion, introducing DPAP into DSNNs and DNNs could effectively improve the accuracy and learning speeds, and extremely compress the networks, showing its general effectiveness, high efficiency and flexible adaptability in various learning tasks and different network structure. More importantly, DPAP, as a brain developmental plasticity-inspired pruning approach, could reveal the brain developmental principle to a certain extent and reflect some characteristics during child development. 

\section{Discussion}
\label{diss}
In this study, we introduced biologically plausible developmental pruning mechanisms into DNNs and SNNs, and demonstrated that the proposed model could help optimize and compress efficient network architectures while improving performance and convergence speed for multiple benchmark datasets. To our best knowledge, this is the first work that studies a purely developmental plasticity-inspired pruning model that brings superior performance while also revealing the naturally occurring pruning processes during brain development. The proposed adaptive pruning strategy is consistent with the developmental pruning mechanism of the brain from multiple perspectives:


\subsubsection{Temporal characteristics capture for SNN} We visualized the process of  temporal spiking sequences acting in DPAP as shown in Fig.\mbox{\ref{discuss}} A. Comparison of the visualized images of spikes and spiking traces reveals that the neuron with a larger number of fired spikes in 20 timesteps correspond to more significant spiking trace. This demonstrates that the spiking trace can represent the activity level of neuron in the temporal dimension. According to the BCM synaptic plasticity theory, our DPAP method not only calculates the current spiking traces of neurons, but also considers the historical mean of neuronal spikes. As shown in Fig.\mbox{\ref{discuss}} A, both synaptic and neuronal pruning importance are calculated relying on the neuronal spiking traces. Hence, our SNN spatial pruning criterions are determined by the unique temporal spiking characteristics.

\subsubsection{Biologically plausible pruning criteria} DPAP method employs local trace-based BCM synaptic plasticity as the measure of synaptic importance. Dendritic spine dynamic plasticity incorporating neuronal activity traces and trace-based BCM synaptic plasticity is used to assess the importance of neurons. Such evaluation criteria are in line with the ``activity-dependent, use it or lose it" developmental principle. Based on these pruning criteria, rarely used and unimportant synapses and neurons are pruned during learning. Fig.\ref{discuss} B,C shows the average importance of all neurons and synapses throughout the learning process. We found that after pruning, the retained neurons and synapses are more important, while relatively unimportant ones were eliminated.

\subsubsection{Biologically plausible gradual decay or even death} 
DPAP prunes synapses or neurons is not an instantaneous impulsive decision, but a continuous and thoughtful evaluation. Pruning occurs after several successive gradual decays, where the survival functions for neurons (Fig.\ref{discuss} D) and synapses (Fig.\ref{discuss} E) gradually decline over a period of time before being pruned, which ensures that the pruned synapses or neurons are redundant. This is also consistent with the biological development that dendritic spine enlargement precedes growth, contraction precedes elimination, and synaptic decay precedes elimination\cite{colman1997alterations}.

\subsubsection{Biologically plausible dynamic pruning} In the brain, pruning is an ongoing process that occurs concurrently with learning\cite{rakic1986concurrent}. The pruning process is not arbitrary, but first drops sharply, then slowly decreases, and finally tends to be stable\cite{huttenlocher1979synaptic}. Specifically, dendritic spine elimination precedes synaptic pruning, and synaptic pruning precedes neural death\cite{trachtenberg2002long,furber1987naturally}. We examined whether the DPAP model can represent these dynamic phenomena in brain developmental pruning. Results are as expected the total number of connections in the network falls sharply at first and then gradually keeps steady during learning (see Fig.\ref{discuss} F). Furthermore, the average number of synapses contained in pruned neurons is 272, while in retained neurons is 298, which indicates that neurons with more pruned synapses are more likely to be deleted. Moreover, the shrinkage and elimination of dendritic spines result from the reduction of neuronal activity and synaptic plasticity, which also leads to the deletion of synapses. Therefore, we can conclude that the shrinkage and elimination of dendritic spines and synaptic pruning are prerequisites for neuronal pruning.

\textbf{Effects of BCM synaptic plasticity} A small number of brain-inspired SNNs pruning methods dynamically prune synapses with smaller weights or decayed weights in shallow SNNs based on the STDP measure of importance\cite{rathi2018stdp,shi2019soft,qi2018jointly,nguyen2021connection}. Although STDP is a feasible and biologically realistic measure of importance, that depends on the spike timing difference of the pre-synaptic and post-synaptic neurons\cite{Mu2008Spike,Bell1997Synaptic,Gerstner1996A}, it is not entirely consistent with the biological LTP and LTD. Unlike STDP and Hebbian\cite{hebb1949first} synaptic modification, BCM theory accounts for experience-dependent synaptic plasticity that could undergo both LTP or LTD depending on the level of post-synaptic response\cite{kirkwood1996experience, bienenstock1982theory}. There is substantial evidence both in the hippocampus and visual cortex that active synapses undergo LTD or LTP depending on the level of post-synaptic spiking which is consistent with the BCM theory\cite{bear1987physiological, abraham2001heterosynaptic, dudek1992homosynaptic, kirkwood1994homosynaptic, artola1990different}. To illustrate the superiority of BCM plasticity, we conducted experiments on SNNs pruning with BCM, STDP, and Hebb plasticity, respectively. As shown in Fig.\ref{discuss} G, replacing BCM with Hebb and STDP will reduce the test accuracy for different datasets, while the more brain-like BCM used in our method achieves the best results.

\section{Conclusion}
\label{conc}
In this paper, we demonstrated that introducing a generalized developmental pruning strategy helped to adaptively compress and optimize the network during the ongoing learning through pruning away redundancy. In contrast to existing pruning methods, our model incorporated multi-scale spatio-temporal developmental plasticity that led to a more compact and efficient network while improving the performance and convergence speed. The proposed method achieves the SOTA results on different datasets for DSNNs. More importantly, our approach shows highly biological plausibility and provides insight into naturally occurring pruning in the developing brain from multiple aspects. In addition, our DPAP algorithm provides a way to achieve the evolution from fixed hierarchical structures to brain-inspired neural circuits, enabling neurons to self-organize to form different neural circuits for performing different tasks. In the future, we expect further studies to combine developmental growth to realize the gradual evolution from small into a complex but efficient network that could adapt to the dynamic changing environment.

\ifCLASSOPTIONcompsoc
  \section*{Acknowledgments}
\else
  \section*{Acknowledgment}
\fi

This work is supported by the Strategic Priority Research Program of the Chinese Academy of Sciences (Grant No. XDB1010302), the National Natural Science Foundation of China (Grant No. 62106261).

\ifCLASSOPTIONcaptionsoff
  \newpage
\fi



%
\bibliographystyle{IEEEtran}
\bibliography{IEEEabrv,elife-sample}

%
\begin{IEEEbiography}[{\includegraphics[width=0.9in,clip,keepaspectratio]{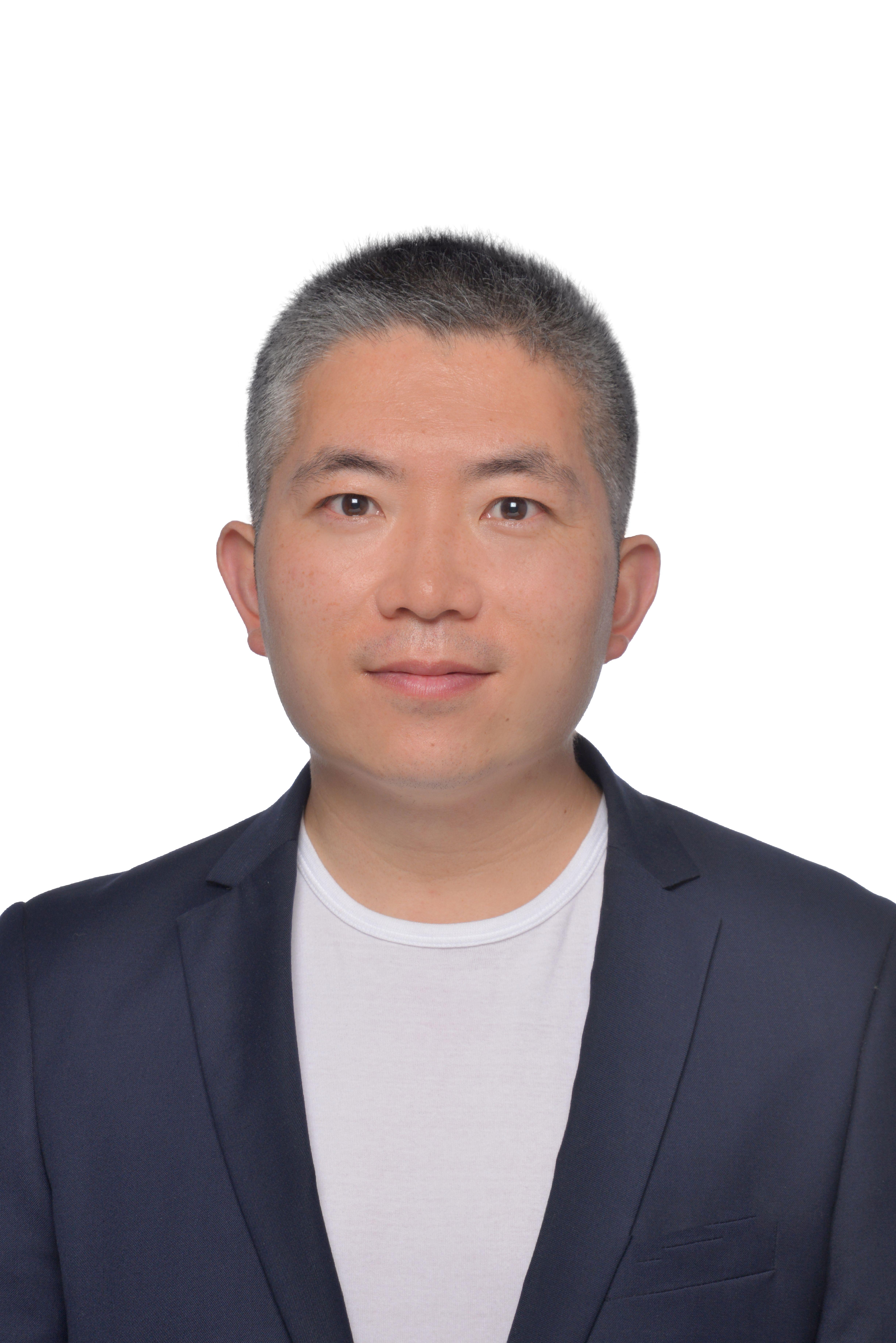}}]{Yi Zeng}
is currently a Professor and Director in the Brain-inspired Cognitive Intelligence Lab, Institute of Automation, Chinese Academy of Sciences (CASIA), China. He is a Principal Investigator in the Key Laboratory of Brain Cognition and Brain-inspired Intelligence Technology, Chinese Academy of Sciences, China, and a Professor in the School of Artificial Intelligence, School of Future Technology, and School of Humanities, University of Chinese Academy of Sciences, China, and a Founding Director of Center for Long-term AI, China. His research interests include brain-inspired Artificial Intelligence, brain-inspired cognitive robotics, ethics and governance of Artificial Intelligence, etc.
\end{IEEEbiography}

\begin{IEEEbiography}[{\includegraphics[width=0.9in,clip,keepaspectratio]{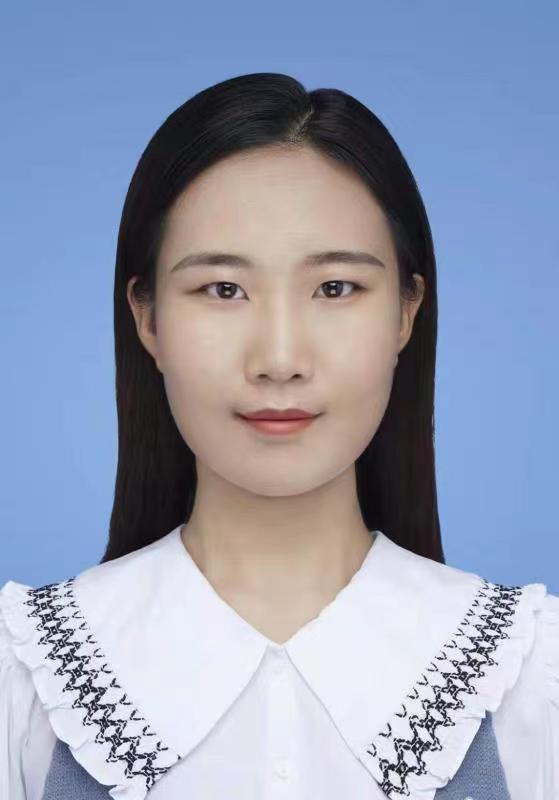}}]{Bing Han}
received the B.Eng. degree in intelligent science and technology from University of Science and Technology Beijing, Beijing, China, in 2021. Now she is the Ph.D. candidate in the Brain-inspired Cognitive Intelligence Lab, Institute of Automation, Chinese Academy of Sciences, supervised by Prof. Yi Zeng. Her current research interests are brain-inspired structural development algorithms for spiking neural networks.
\end{IEEEbiography}

\begin{IEEEbiography}[{\includegraphics[width=0.9in,clip,keepaspectratio]{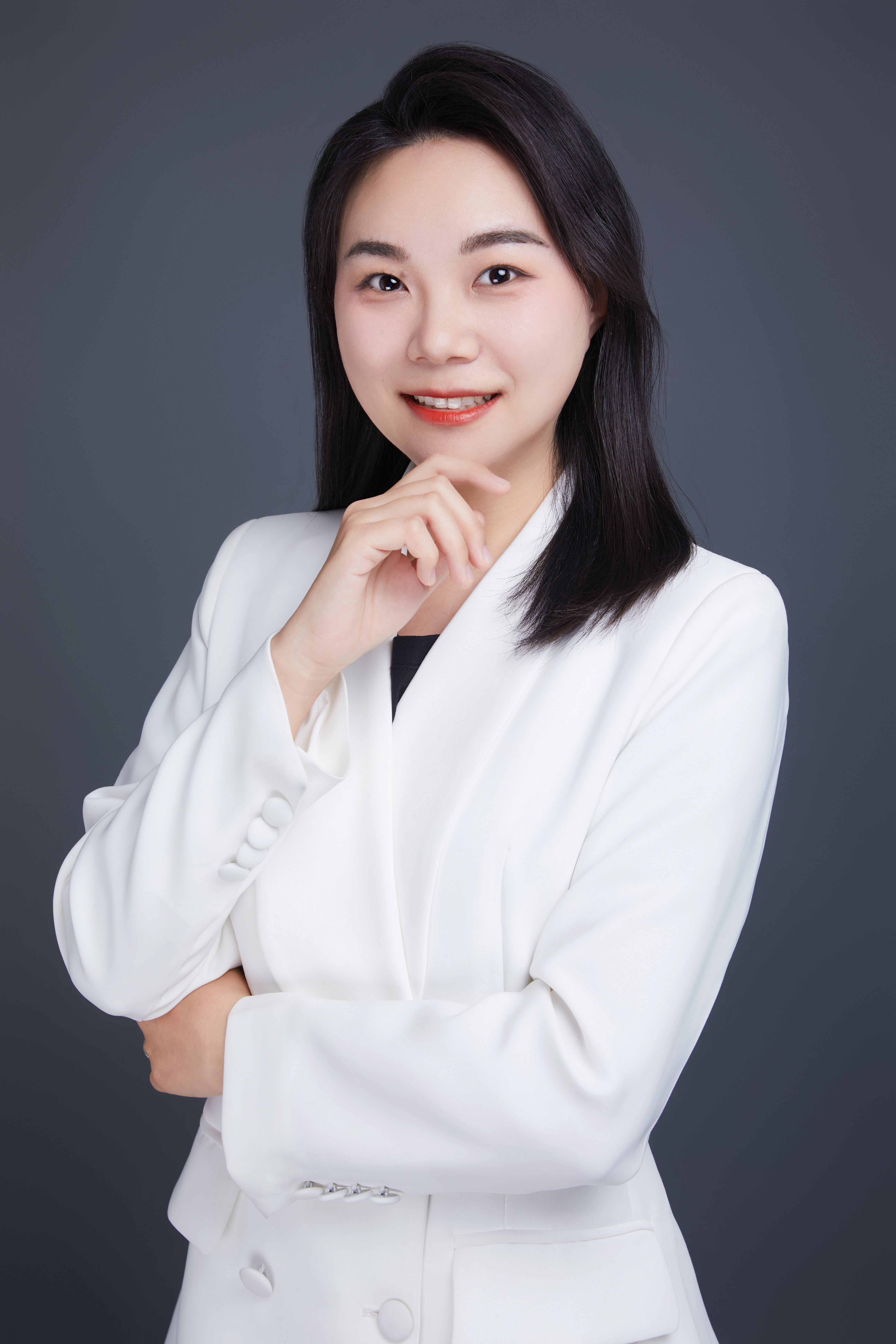}}]{Feifei Zhao}
is currently an Associate Professor in the Brain-inspired Cognitive
Intelligence Lab, Institute of Automation, Chinese Academy of Sciences(CASIA), China.
She received the Ph.D. degree from the University of Chinese Academy of Sciences, Beijing, China, in 2019. Her current research interests include Brain-inspired Developmental and Evolutionary Spiking Neural Networks, Empathy driven AI Ethics and Safety.
\end{IEEEbiography}

\begin{IEEEbiography}[{\includegraphics[width=0.9in,clip,keepaspectratio]{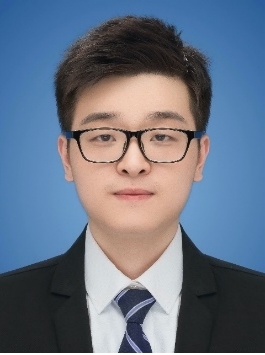}}]{Guobin Shen}
	received his B.Eng. degree from Sun Yat-sen University in Guangzhou, Guangdong, China. He is now a PhD candidate in the Brain-inspired Cognitive Intelligence Lab, at the Institute of Automation, Chinese Academy of Sciences, under the supervision of Prof. Yi Zeng. His research focuses on biologically-inspired learning algorithms and spiking neural network architecture design and training strategies.
\end{IEEEbiography}




\end{document}